# A Multi-Transformation Evolutionary Framework for Influence Maximization in Social Networks

Chao Wang, Jiaxuan Zhao, Lingling Li, Licheng Jiao, Jing Liu, and Kai Wu, Xidian University, CHINA

*Abstract*—Influence maximization is a crucial issue for mining the deep information of social networks, which aims to select a seed set from the network to maximize the number of influenced nodes. To evaluate the influence spread of a seed set efficiently, existing studies have proposed transformations with lower computational costs to replace the expensive Monte Carlo simulation process. These alternate transformations, based on network prior knowledge, induce different search behaviors with similar characteristics to various perspectives. Specifically, it is difficult for users to determine a suitable transformation a priori. This article proposes a multi-transformation evolutionary framework for influence maximization (MTEFIM) with convergence guarantees to exploit the potential similarities and unique advantages of alternate transformations and to avoid users manually determining the most suitable one. In MTEFIM, multiple transformations are optimized simultaneously as multiple tasks. Each transformation is assigned an evolutionary solver. Three major components of MTEFIM are conducted via: 1) estimating the potential relationship across transformations based on the degree of overlap across individuals of different populations, 2) transferring individuals across populations adaptively according to the inter-transformation relationship, and 3) selecting the final output seed set containing all the transformation's knowledge. The effectiveness of MTEFIM is validated on both benchmarks and real-world social networks. The experimental results show that MTEFIM can efficiently utilize the potentially transferable knowledge across multiple transformations to achieve highly competitive performance compared to several popular IM-specific methods. The implementation of MTEFIM can be accessed through the external link in the footnote[1].

## I. Introduction

Social networks are employed to describe the communication in convention and connection between people in a word-of-mouth manner [1]. With the rapid increase in the number of users, new online social network sites or apps such as Twitter, MicroBlog, TikTok, and Instagram have become the mainstream networking platforms for people to share and exchange ideas, providing great convenience for marketers to target prospective customers [2]-[3].

In recent years, the rise of these websites or apps has boosted a complete industrial chain in social media marketing, including influencers and marketing agencies [4]. Influencers, such as Charli D' Amelio and Addison Rae, have tens of millions of followers on TikTok. Usually, these influencer marketing agencies promote their products by collaborating with initial influencers (such as actors and celebrities), which is proven effective and successful by the market. Finding influencers who can help engage new consumers exponentially is essential to the success of influencer marketing [5]. Coincidentally, this issue also appears in a series of practical applications in political campaigns [6] and epidemic analyses [7], which is at the core of the influence maximization (IM) problem.

The above IM problem intends to select a group of users from social networks, called a seed set, to maximize the spread of influence under an information diffusion process [8]. Some popular models have been proposed to study the diffusion process, such as the independent cascade (IC) [9], weighted cascade (WC) [10], and linear threshold (LT) [11], which simulate the independent or collective behavior of the seed set. Since information production is interfered by a variety of social or biological factors in different problems, existing works tend to employ network structure information to find those influencers under these models [12]. Domingos *et al*. [8] first formalized IM as an NP-hard combinatorial optimization problem, which shows that only the optimal set with a certain degree of approximation could be found. Domingos *et al*. [8] further proposed a series of greedy-based IM algorithms.

This work was supported in part by the Key Scientific Technological Innovation Research Project by Ministry of Education, the National Natural Science Foundation of China Innovation Research Group Fund(61621005), the State Key Program and the Foundation for Innovative Research Groups of the National Natural Science Foundation of China (61836009), the Major Research Plan of the National Natural Science Foundation of China (91438201, 91438103, and 91838303), the National Natural Science Foundation of China (U1701267, 62076192, 62006177, 61902298, 61573267, and 61906150), the 111 Project, the Program for Cheung Kong Scholars and Innovative Research Team in University (IRT 15R53), the ST Innovation Project from the Chinese Ministry of Education, the Key Research and Development Program in Shaanxi Province of China(2019ZDLGY03-06), the National Science Basic Research Plan in Shaanxi Province of China(2019JQ-659, 2022JQ-607), the Scientific Research Project of Education Department In Shaanxi Province of China (No.20JY023), the fundamental research funds for the central universities (XJS201901, XJS201903, JBF201905, JB211908), and the CAAI-Huawei MindSpore Open Fund. Corresponding author: Licheng Jiao (e-mail: lchjiao@mail.xidian.edu.cn).

[1] https://github.com/xiaofangxd/MTEFIM

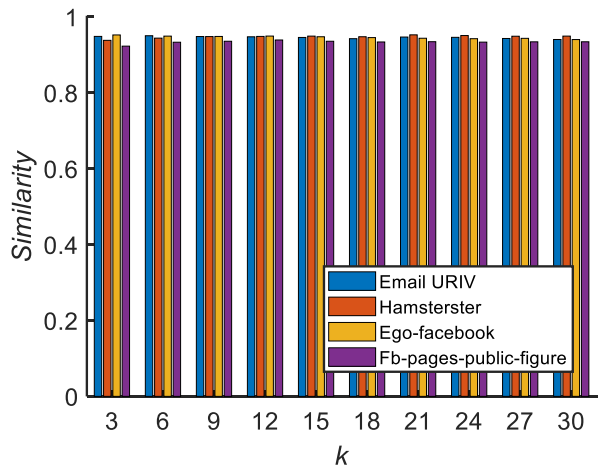

Fig. 1. The similarity of two optimization tasks, namely, the optimizing of EDV and the optimizing of TIS, on four real-world networks: Email URIV, Hamsterster, Ego-facebook, and Fb-pages-public-figure, where $K$ is the size of the seed set.

Theoretical results show that under the IC model, the greedy method can produce an optimal solution with an accuracy of no less than $1-1/e$. Then, several algorithms are developed to address the inefficiency and insufficient scalability of the greedy method. Leskovec *et al*. [13] proposed an improved greedy algorithm with a "Cost-Efficient Lazy Forward (CELF)" strategy, which is reported to be 700 times faster than the greedy method. Goyal *et al*. [14] further extended CELF++ by considering the submodularity property of IM. In addition, Chen *et al*. [10] proposed a new greedy algorithm to conduct the diffusion process on a network after pruning. However, these simulation-based greedy methods require thousands of Monte Carlo simulations to evaluate influence spread, which is unattainable in practice.

To overcome the limitations of simulation-based methods, several heuristic methods have been presented to handle large-scale networks. A simple ranking idea can be employed to select the seed set, such as using degree, PageRank [15], and distance centrality [16]. Chen *et al*. [10] designed a degree discount heuristic method, which assumes that the influence spread increases with the user's degree. Wang *et al*. [17] further extended to a generalized degree discount heuristic. However, these approximation methods may be quite different from the influence diffusion process (see the experimental analysis in [9]-[10]).

In the last decade, a series of meta-heuristic algorithms have been proposed to solve IM problems due to their superiority in solving NP-hard problems, such as the simulated annealing (SA) and the evolutionary algorithm (EA). These methods have excellent performance on many real-world networks. Jiang *et al*. [18] proposed a cheap approximate model, named the expected diffusion value (EDV), to replace the diffusion process and optimize the objective by the SA algorithm. Lee *et al*. [19] introduced an approximation model for the diffusion process by limiting the spread on users with two hops of the seed set. Gong *et al*. [20] reduced the search space for IM problems by using community information in a social network, and proposed a memetic algorithm to optimize the two-hop influence spread, called TIS in this article. In addition, Gong *et al*. [21] proposed a discrete particle swarm optimization algorithm to optimize the local influence estimation (LIE) that approximates the 2-hop influence spread. Wang *et al*. [22] presented a new influence estimation model by considering the estimate and variance of users in a 2-hop neighborhood, called IEEV in this article. Singh *et al*. [23] introduced a learning-based particle swarm optimization for IM and extended the EDV to approximate two-desired areas. In addition, an ant colony algorithm [24] is proposed to optimize the local influence evaluation. Li *et al*. [25] presented a discrete crow search algorithm based on the network topology structure. Ma *et al*. [26] proposed a novel evolutionary deep reinforcement learning algorithm for IM problems in complex networks, where the IM problem is transformed into a continuous weight optimization problem of deep Q-networks. Furthermore, many methods combining reinforcement learning and network embedding have been proposed to optimize the expected influence to solve various IM problems [27]-[28]. These methods construct a variety of cheap proxy models for influence spread from different perspectives and employ the meta-heuristic paradigm to optimize them. However, the similarities between these proxy models have not been investigated, which may improve the performance of the IM algorithm. In addition, the work [29] pointed out that these proxy models induce different search behaviors with similar characteristics. Therefore, it is difficult to choose the most appropriate proxy model when dealing with actual IM problems.

The similarity between optimization tasks can be reflected by an overall correlation among the fitness landscapes, which can be quantified by Spearman's rank correlation [30]. One million seed sets are randomly generated in the search space with a fixed number of seeds, and their valuations are calculated on two popular proxy models, EDV and TIS. Then, the Spearman's rank correlation of their valuations is regarded as the similarity of two optimization tasks [31], namely, the optimizing of EDV and the optimizing of TIS. Fig. 1 shows the similarity of the two optimization tasks on four representative real-world networks. The fitness landscapes of these two optimization tasks are indeed highly similar because these proxy models all consider the neighbor information of the seed set, leading to a large amount of available knowledge between the optimization processes of the different proxy models.

Multi-transformation optimization (MTFO) [32] refers to the simultaneous optimization of multiple alternative formulas for a target optimization task, which can be solved by multi-task optimization (MTO) methods [33]. Due to the implicit parallelism of EAs, multi-task optimization has attracted significant attention in the field of evolutionary computation (EC) [33]- [37]. MTFOs solved by multi-task EAs have been applied to many practical problems, such as high-dimensional optimization [38]-[39] and expensive optimization [40]-[41]. Compared with a single-transformation EA optimizing a certain formula of the target task alone, a multi-transformation EA can utilize the unique advantages of different formulas to significantly improve performance through knowledge transfer in the evolutionary process. Inspired by the multi-transformation EA, this article presents a multi-transformation evolutionary framework for IM (MTEFIM) to utilize the similarities and unique advantages of multiple proxy models to improve the performance of evolutionary-based IM algorithms and to prevent users from choosing proxy models a priori. Multiple proxy models, called multiple transformations in this article, are optimized simultaneously. Each transformation is assigned a population. The degree of overlap between individuals of different transformations provides an indirect estimate of their potential relationships. Based on the inter-transformation relationship, a knowledge transfer process across a given transformation and the most relevant "assisted" transformation is designed to exchange common information adaptively. Finally, MTEFIM considers the comprehensive rank on the optimal seed set of each transformation and then outputs a final seed set containing the knowledge of all proxy models. Empirical studies are conducted on a series of synthetic benchmarks and real-world networks to validate the performance of MTEFIM. The results show that MTEFIM achieves a highly competitive performance in terms of the influence spread and the running time by knowledge transfer across transformations compared to several popular IM methods.

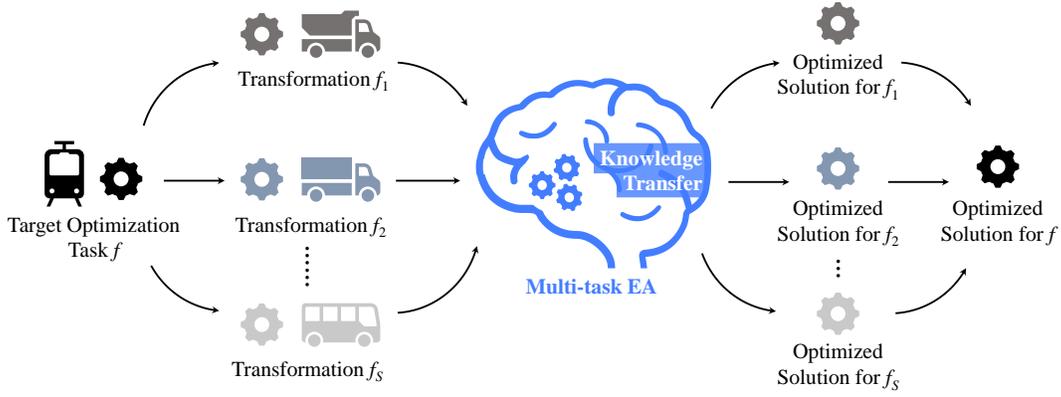

Fig. 2. The core idea of MTFO, where multiple alternate transformations of a target task are optimized by multi-task EA simultaneously.

The main research contributions of this article are presented as follows:

1) Inspired by the similarities across transformations, a multi-transformation evolutionary framework for IM with convergence guarantees is proposed to implicitly utilize the common and unique knowledge of multiple transformations. Apart from the existing methods that optimize one proxy model alone, the proposed method optimizes multiple transformations simultaneously in one run to avoid transformation selection a priori.

2) By considering the degree of overlap between individuals of different transformations, a novel inter-transformation relationship estimation strategy is presented to guide the adaptive knowledge transfer process in a multi-transformation environment. To yield the final output seed set, a simple yet effective selection strategy is introduced, which considers compromise performance on the optimal seed set of each transformation.

The remainder of this article is organized as follows. Section II provides background knowledge on IM problems, proxy models, and MTFO. MTEFIM is introduced in detail in Section III. Section IV describes the experimental results and discussion on five real-world networks. Finally, conclusions and future works are presented in Section V.

## II. PRELIMINARY

In this section, background knowledge on IM in social networks and proxy models for meta-heuristic methods are introduced first. Then, a brief review of the basics and related work in the field of MTFO is provided.

### A. IM in Social Networks

A social network is modeled as a graph $G = (V, E)$, where $V = \{1, 2, …, v\}$ and $E = \{e_{ij} \mid i, j \in V\}$ are accordingly the node-set and edge set of the network[42]. Given a seed set $A \in V$ with $k$ seed nodes, a probabilistic influence spread model returns the expected number of nodes influenced by the seeds, which is denoted as $\sigma(A)$. Many models are proposed to describe the influence spread process under different scenarios, such as the IC, WC, and LT models. The IC model is one of the most widely studied diffusion models, and many proxy models are developed to approximate this model [43]-[44]. Therefore, this article chooses the IC model to describe the influence spread process.

In the IC model [9], all the nodes are either active or inactive. Activated nodes influence inactive nodes through connections between nodes with a propagation probability $p$. Considering a seed set $A \in V$ with $k$ nodes, the influence spread process can be expressed as follows. In the first step, $k$ seeds in $A$ are active and saved in set $A_1$. At step $t$, given the propagation probability $p(u, v)$ of the edge $e_{uv}$, a node $u \in A_{t-1}$ may activate an inactive neighbor $v$ with the probability $p$. Note that $u$ only has one chance to activate its neighbors. Those successfully activated nodes are saved in the set $A_t$. The above influence spread process stops when $A_t = \varnothing$. Then, the number of active nodes in the influence spread process is defined as the influence spread $\sigma(A)$ of seed set $A$ under the IC model.

IM can be considered as the following discrete problem [45]: Given a social network $G = (V, E)$ and the size of seed set $k$, the goal of the IM problem is to select a seed set $A$ under an influence spread model such that,

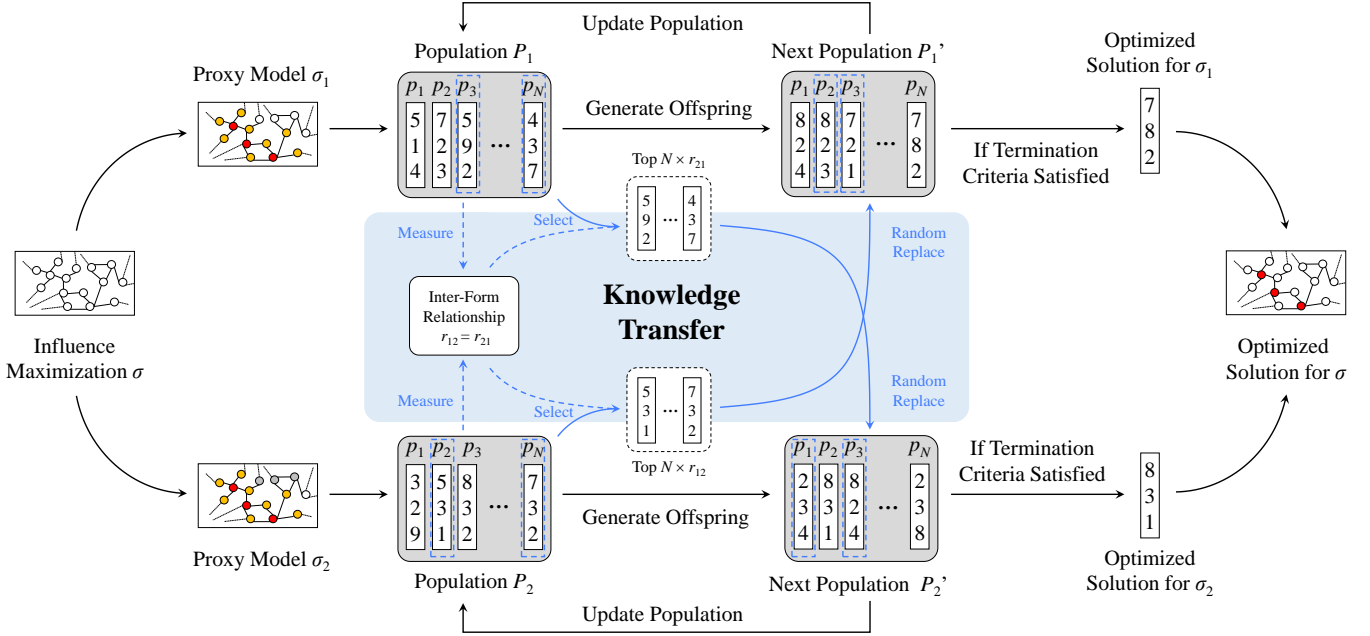

Fig. 3. The outline of MTEFIM for bi-transformation IM.

$$\arg\max_{A \subseteq V} \sigma(A) \ s.t. |A|=k \tag{1}$$

where $|A|$ is the size of set $A$. To solve the above IM problem, tens of thousands of Monte Carlo simulations are used to estimate the influence spread process of any given seed set, which is very time-consuming [43]. In the next section, two popular proxy models, EDV and TIS, are introduced to approximate influence spread $\sigma$ to replace the expensive Monte Carlo simulation process.

B. *Proxy Models for Meta-heuristic Methods*

**EDV** [18]. For a small propagation probability $p$ in the IC model, EDV estimates the expected number of nodes influenced by seed set $A$ with $k$ seeds through,

$$EDV(A)=k+\sum_{b \in NB(A)\setminus A}\left(1-(1-p)^{\delta(b)}\right), \tag{2}$$

where $NB(A)=A\cup\{b|\exists a\in A, e_{ab}\in E\}$ represents one-hop neighbors of the seed set $A$, and $\delta(b)=|\{a|a\in A, e_{ab}\in E\}|$ represents the influence from $A$ to $b$. $|*|$ is the number of elements in the set $*$. This formula essentially extracts the edge and node information of $N(A)$ to form a subgraph to quickly evaluate the influence spread. A series of existing efforts use meta-heuristic algorithms to optimize EDV or its variants, such as the SA [18], the particle swarm optimization (PSO) [23], [46], and the ant colony optimization [24]. Compared with heuristic-based methods, these methods can quickly find more accurate solutions in real-world networks.

**TIS** [20]. For the IC model, TIS estimates the influence spread of the seed set A by the 2-hop neighborhood through,

$$TIS(A)=\sum_{a\in A}TIS\{a\}$$
$$-\left(\sum_{a\in A}\sum_{b\in N(a)\cap A}p(a,b)(\alpha(b)-p(b,a))\right)-\beta, \tag{3}$$
$$\beta=\sum_{a\in A}\sum_{b\in N(a)/A}\sum_{c\in N(b)\cap A/\{a\}}p(a,b)p(b,c)$$

where $N(*)$ denotes the 1-hop nodes cover of seed $*$, $p(a, b)$ represents the propagation probability between the active node $a$ and the inactive node $b$, and $\alpha(b)$ represents the 1-hop influence spread of the node $b$. In (3), the first term evaluates the sum of the 2-hop influence spread of seeds in $A$. The second and third terms consider the potential redundant influences between the seeds. TIS considers the information of the seed neighbors and the neighbors' neighbors to quickly approximate the influence spread of the seed set, which has been widely extended to solve IM problems on different types of networks [25], [47]. Based on TIS, some proxy models are proposed, such as LIE [21], [48], and IEEV [22].

Since different proxy models consider different neighbor information on the seed set, the use of EDV and TIS may lead to different search behaviors with similar characteristics, thus resulting in their different performances on specific problems. The EDV and the TIS have been emphasized in previous studies, so this article chooses them as alternative transformations of IM problems. Furthermore, in Section V. E, more comparative experiments are provided to illustrate the rationale for choosing EDV and TIS.

C. *MTFO*

Given $S$ alternate formulations $\{f_1, f_2, …, f_S\}$ of a target minimization task $f$, the MTFO problem can be formalized as [32],

$$\begin{aligned}\min_{X} f_i(x), i &= 1,2,...,S \\ s.t.\ x &= \left[x^1, x^2,...,x^m\right]\in D\end{aligned}, \tag{4}$$

where $D$ is the search space for the target optimization task $f$ and $m$ is the dimension of $x$. The purpose of MTFO is to find the best solutions to an optimization task $f$ through effective knowledge transfer across alternate formulations. This solution can be selected from the following set,

$$X^* = \{x^* | x^* = \arg\min f_i(x), i=1,2,...,S\}. \tag{5}$$

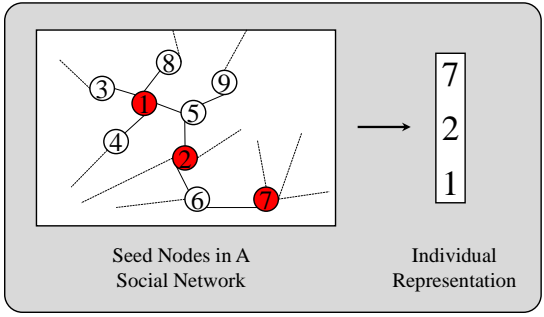

Fig. 4. An illustration of the representation in a social network with 9 nodes, where $k$=3.

In general, the MTFO problem is regarded as an MTO problem. Evolutionary multi-task optimization (EMTO) aims to solve multiple tasks simultaneously by utilizing some paradigms motivated by EC [49]-[50]. Gupta et al. [33] designed the multifactorial EA (MFEA) to optimize many tasks at the same time, inspired by cultural exchanges between organisms. MFEA encodes individuals into a unified search space to provide a basic guarantee for knowledge transfer and sets a skill factor for each individual to indicate its associated task. By assortative mating and selective imitation, knowledge is implicitly transferred across tasks with a certain probability *rmp* to improve the convergence performance. A theoretical analysis of MFEA demonstrates that knowledge transfer across tasks is controlled by *rmp*. Bail et al. [36] further proposed MFEAII, based on online parameter estimation, to address the prevalence of negative transfers. An explicit EMTO algorithm, called EMEA in this article, was proposed by Feng et al. [34] to utilize multiple evolutionary solvers with different biases. In EMEA, each task is assigned an evolutionary solver, and the top individuals on each task are explicitly transferred to other tasks based on an autoencoding that learns optimal linear mapping across tasks. Liaw et al. [35] proposed symbiosis in biocoenosis (SB) optimization to deal with MTO, which leverages the concept of SB to transfer knowledge across tasks. To the best of our knowledge, although a series of excellent EMTO methods have been presented, none is specifically designed to address the IM problem.

Fig. 2 illustrates the core idea of the MTFO, where multiple alternate transformations of a target task are optimized by a multi-task EA at the same time. The unique advantages and similarities of the alternative formulas are exploited through the knowledge transfer that occurs in the search process. The basic idea of MTFO is employed to solve complex optimization problems in existing works [38]-[41]. Ding et al. [40] proposed an effective knowledge transfer across low-fidelity and high-fidelity transformations for a single expensive problem. To solve large-scale optimization problems, existing works use dimensionality reduction methods to construct low-dimensional optimization transformations, such as the unsupervised neural network and the random embedding [38]. Feng et al. [39] proposed a multi-variation search to optimize high-dimensional and low-dimensional transformations simultaneously. Multi-variation collaborative work not only utilizes the unique advantages of each alternative transformation, but also avoids transformation selection for

**Algorithm 1: Framework of MTEFIM**
**Input**:
   $\sigma=\{\sigma_i, 1 \leq i \leq S\}$: $S$ proxy models (transformations) for IM under an influence spread model;
   $N$: population size of each transformation;
   $k$: size of seed set;
   *MFE*: maximum number of function evaluations;
**Output**:
   $A^*$: output solution for IM.

1. $P=\{P_i, 1 \leq i \leq S\}\leftarrow$Initialize $S$ populations;
2. $\varphi_P=\{\varphi_{Pi}, 1 \leq i \leq S\}\leftarrow$Evaluate $S$ populations on their transformations;
3. **while** (stopping criterion is not met) **do**
4.   $R\leftarrow$Learn the inter-transformation relationship according to $P$ (see **Algorithm 2**);
5.   **for** $i$=1 **to** $S$ **do**
6.     $O_i\leftarrow$Generate offspring based on designed genetic operator;
7.   **end for**;
8.   **for** $i$=1 **to** $S$ **do**
9.     **if** (stopping criterion of $\sigma_i$ is not met) **then**
10.       $p\leftarrow \arg\max_{p\in\{1,2,...S, p\neq i\}} r_{i,p}$;
11.       **if** rand(0, 1) $< r_{i,p}$ **then**
12.         $O_i\leftarrow$Select $N \cdot r_{i,p}$ individuals from $O_p$ to replace the individuals in $O_i$ randomly;
13.       **end if**
14.       $\varphi_{Oi}\leftarrow$Evaluate $O_i$ on transformation $\sigma_i$;
15.       $P_i\leftarrow$Select top $N$ individuals from $P_i \cup O_i$ based on fitness value $\varphi_{Pi} \cup \varphi_{Oi}$;
16.     **end if**;
17.   **end for**;
18. **end while**;
19. $A^*\leftarrow$Select output solution from $P$ by (9).

users a priori. However, due to the specificity of individual representation for IM problems, the existing MTFO methods cannot directly deal with a multi-transformation IM.

### III. THE PROBLEM FORM AND MTEFIM

#### A. Outline of MTEFIM

Given $S$ proxy models $\{\sigma_1, \sigma_2, ..., \sigma_S\}$ of the IM problem $\sigma$ on a social network $G = (V, E)$ under an influence spread model, the multi-transformation IM can be expressed as,

$$\max_{A \subseteq V} \sigma_i(A), i = 1, 2, ..., S \\ s.t. |A| = k \qquad (6)$$

where the size of the seed set selected from the network is constrained to $k$. All transformations share the same search space. The intention is to optimize all transformations simultaneously, exploiting their potential knowledge by evolutionary multitasking.

The outline of MTEFIM for a bi-transformation IM is shown in Fig. 3. As observed, each transformation is assigned a population. The individuals, i.e., seed sets, in all populations are initialized in the same search space $A \in V$. Each population is only evaluated on the corresponding transformation. In the main loop, for any two transformations $\sigma_i$ and $\sigma_j$, the inter-transformation relationship $r_{ij}$ is estimated by the degree of overlap across individuals of different transformations online. Each population is employed to generate the offspring population based on the designed genetic operator. Then, for each transformation, based on the inter-transformation relationship $R=\{r_{ij}|1 \leq i, j \leq S, i \neq j, r_{ij}=r_{ji}\}$, the top individuals from the most related transformation are transferred to randomly replace some individuals in the current transformation, as shown in Fig. 3, where the transferred individuals are viewed as the carriers of common knowledge across transformations. Then, the next population from the parent and offspring populations are selected by elitism. The final output solution is selected from the optimal solution of each transformation, which contains all the proxy model knowledge. In MTEFIM, each transformation is distributed with an evolutionary solver. Individuals that may carry shared knowledge flow across transformations. The framework of MTEFIM for multi-transformation IM is described in Algorithm 1. Next, the details of MTEFIM are given in this section.

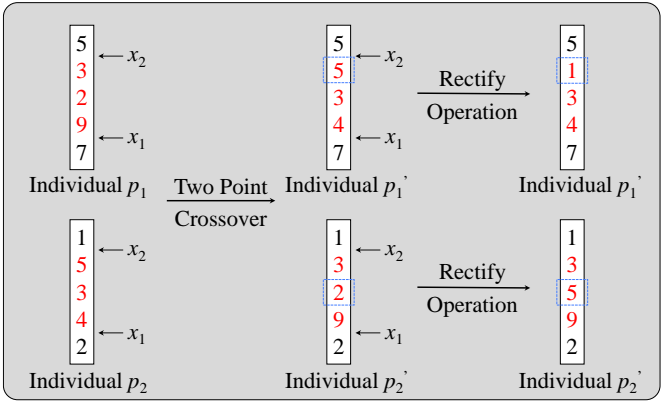

Fig. 5. An illustration of the two-point crossover on individual $p_1$ and $p_2$.

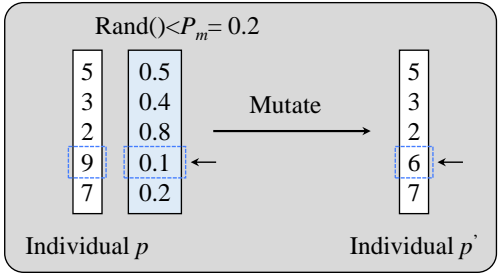

Fig. 6. An illustration of the mutation operator on individual $p$.

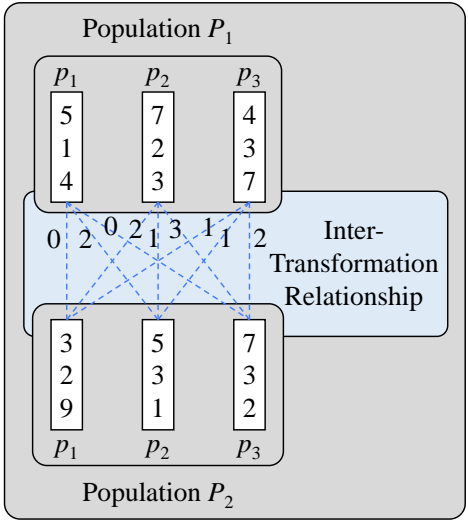

Fig. 7. An illustration of the inter-transformation relationship estimation across two transformations, where inter-transformation relationship $r_{12} = r_{21} = \frac{0+2+0+2+1+3+1+1+2}{(3\times 3 \times 3)} \approx 0.44$.

### B. Population Initialization

In MTEFIM, each individual $p$ in all populations represents a seed set with $k$ nodes, which can be indicated as,

$$p = \{p^1, p^2, ..., p^k\}, \quad (7)$$

where $p^k$ is the node index in $\{1, 2, …, v\}$, and $v$ is the number of nodes in the social network $G$. Fig. 4 shows an illustration of the representation.

Since random initialization often leads to a slower convergence speed, MTEFIM utilizes a warm-starting method commonly used in IM, that is, the degree discount heuristic [21]. It is worth noting that other heuristics are still applicable. The degree heuristic-based warm starting method is described as follows:

1) Each individual is initialized by selecting $k$ seeds with the highest degree in the social network $G$;

2) Each node contained in each individual is replaced with a certain probability of 0.5 by its neighbor nodes in the social network except the nodes are contained in the current individual.

After performing the above initialization, a population with degree-based knowledge and diversity can be obtained.

### C. Genetic Operators

After the population is initialized, genetic operators including crossover and mutation are performed on the initial population to generate a better seed set.

For every two individuals in the population, the seed information is shared to obtain a higher potential seed set in the crossover operator with the probability $p_c$. In this article, the two-point crossover [51] is employed due to its high efficiency and low complexity. An illustration of the two-point crossover on individuals $p_1$ and $p_2$ is shown in Fig. 5. Given two parent individuals $p_1$ and $p_2$ from the parent population, two crossover positions $x_1$ and $x_2$ are generated first, where $\{x_1, x_2|1 \leq x_1, x_2 \leq k, x_1 < x_2\}$. Then, two new candidates, $p_1'$ and $p_2'$, are generated by swapping the seed between positions $x_1$ and $x_2$ of $p_1$ and $p_2$. Next, the validity of $p_1'$ and $p_2'$ should be guaranteed, that is, there is no identical seed in an individual. Specifically, new seeds are randomly generated to replace repetitive seeds in candidates.

For each individual $p$ in the population, the following mutation operator [52] is conducted to help escape from local optima after crossover. An illustration of the mutation operator on individual $p$ is given in Fig. 6. Each seed in the individual is replaced with other nodes in the network with a certain probability $p_m$ and ensure the feasibility of the individual.

**Algorithm 2: Inter-transformation Relationship Estimation**

**Input**:
   $P=\{P_i, 1 \leq i \leq S\}$: $S$ populations for $S$ transformations;
   $N$: population size of each transformation;
   $k$: size of seed set;

**Output**:
   $R$: inter-transformation relationship.

1.    $R \leftarrow \{r_{ij}=0 | 1 \leq i, j \leq S, i \neq j\}$;
2.    **for** $i=1$ **to** $S$ **do**
3.      **for** $j=i+1$ **to** $S$ **do**
4.        **for** each individual $p'$ in $P_i$ **do**
5.          **for** each individual $p''$ in $P_j$ **do**
6.            $r_{ij} \leftarrow r_{ij}$ + Count the number of identical seeds in two seed sets $p'$ and $p''$;
7.          **end for**
8.        **end for**
9.        $r_{ij} \leftarrow r_{ij} / (k \times N \times N)$;
10.       $r_{ji} \leftarrow r_{ij}$;
11.      **end for**
12.   **end for**

*D. Estimating the Inter-Transformation Relationship*

The inter-transformation relationship is employed to guide the knowledge transfer across different transformations. During the evolution, each transformation is assigned a selected most relevant auxiliary transformation. Knowledge transfer takes place between them with the individual as a carrier. Generally, the more similar the two transformations are, the more common knowledge they may contain. This is because similar transformations contain more common knowledge for searching, for example, similar fitness landscapes (see Fig. 1). Because landscape analysis is computationally expensive, populations are used to implicitly estimate the inter-transformation relationship in the multi-transformation environment.

The degree of overlap between two seed sets can directly reflect their correlation. Therefore, the degree of overlap between individuals in the two populations corresponding to two transformations is estimated to obtain the inter-transformation relationship $r_{ij}$. The inter-transformation relationship estimation is shown in Algorithm 2. The proportion of the same seeds in the corresponding seed sets of the two populations is calculated to approximate the relationship across transformations. Fig. 7 demonstrates the inter-transformation relationship estimation across two transformations.

If the computational complexity of Step 6 in Algorithm 2 is $O(k^2)$, the computational complexity of the inter-transformation relationship estimation is,

$$O\left(\frac{S(S-1)}{2} \cdot \frac{N(N-1)}{2} \cdot k^2\right) = O(S^2 N^2 k^2), \quad (8)$$

where $S$, $N$, and $k$ are accordingly the number of transformations, population size and seed set size.

*E. Selecting Output Seed Set*

Since each transformation addresses the IM problem from a different perspective, their optimal seed sets may also differ. An intuitive idea is to use Monte Carlo simulation to evaluate each optimal seed set and select the best-performing seed set as the output seed set, named MCSS, but it is costly due to multiple simulations.

After obtaining the optimal individual of each transformation, an easy-to-implement method for selecting an output seed set (SOSS) is proposed. Given $S$ optimal individuals $p^* = \{p_1^*, p_2^*, \ldots, p_s^*\}$ for $S$ transformations $\sigma = \{\sigma_i, 1 \leq i \leq S\}$, this can be specified as follows:

1) Each optimal individual $p_i^*$ is evaluated on all transformations $\sigma$.

2) The rank index $rank_{ij}$ of each optimal individual $p_i^*$ on the ascending factor cost list is calculated corresponding to each transformation $\sigma_j$.

3) Sum the rank index $rank_{ij}$ of each optimal individual $p_i^*$ by,

$$CR = \sum_j C_j rank_{ij}, \quad (9)$$

where $C_j$ represents the user's prior preference for the $j$-th transformation and meets the following condition,

$$\sum_{j=1}^{S} C_j = 1, 0 < C_j \leq 1. \quad (10)$$

The user's prior preference $C_j$ can be set to $1/S$ when not provided by the user. This means that there is no obvious determinable preference between the transformations.

4) The seed set with the top cumulative rank $CR$ is employed as the final output seed set.

This method comprehensively considers the performance of the optimal seed set in all transformations and avoids an expensive Monte Carlo simulation process.

*F. Theoretical Analysis of MTEFIM*

In this section, an asymptotic global convergence analysis of MTEFIM is presented. Without loss of generality, the sub-population and objective function associated with the transformation $s$ are defined as $P_s$ and $\sigma_s$, respectively. At time step $t$, the $s$-th sub-population is assumed to obey the probability distribution $p^s(x,t)$. Thus, in the $t$-th iteration of MFEAIM, the offspring population generated for the $s$-th transformation is considered to be drawn from the following mixed probability distribution.

**Lemma 1.** *Assuming that a parent-centered genetic operator is employed in the MTEFIM, the mixed probability distribution $p^s(x,t)$ of the offspring for the s-th transformation at time step $t$ can be expressed as follows,*

$$p^s(x,t) = \left[1 - r_{s,j^*}^2(t)\right] \cdot p^s(x,t) + r_{s,j^*}^2(t) \cdot p^{j^*}(x,t)$$

$$j^* = \underset{j \in \{1,2,\ldots,s\}, j \neq s}{\arg\max} r_{s,j}(t) = \frac{\sum_{m=1}^{N}\sum_{n=1}^{N} I(x_s^m(t), x_j^n(t))}{N \times N \times K}. \quad (11)$$

*Proof.* In a multi-transformation environment, an offspring solution $x$ in sub-population $s$ is drawn from either the probability distribution $p^s(x,t)$ or $p^{j^*}(x,t)$. The probability $P(x \sim p^s(x,t) | x \in s)$ that $x$ is drawn from $p^s(x,t)$ occurs under two cases

TABLE I
THE BASIC CHARACTERISTICS OF FIVE REAL-WORLD NETWORKS USED IN THIS ARTICLE.

| Networks | /V/ | /E/ | AC | p |
|---|---|---|---|---|
| Email URIV | 1.1K | 5.5K | 0.220 | 0.05 |
| Hamsterster | 2.4K | 16.6K | 0.538 | 0.03 |
| Ego-facebook | 4.0K | 88K | 0.605 | 0.02 |
| Fb-pages-public-figure | 11.6K | 67K | 0.179 | 0.04 |
| NetHEPT | 15.2K | 58.9K | 0.346 | 0.05 |

(see Algorithm 1):

Case 1: When a random number $rand()$ is greater than or equal to $r_{s,j^*}(t)$, all individuals of the offspring $x \in s$ are derived from $p^s(x,t)$. The probability of the case is,

$$P(case1) = 1 - r_{s,j^*}(t). \quad (12)$$

Case 2: When a random number $rand()$ is less than $r_{s,j^*}(t)$, the probability that the offspring $x \in s$ is derived from $p^s(x,t)$ is,

$$P(case2) = r_{s,j^*}(t)\left(1 - r_{s,j^*}(t)\right). \quad (13)$$

Therefore, the probability $P(x \sim p^s(x,t)|x \in s)$ can be expressed as,

$$P\left(x \sim p^s(x,t) | x \in s\right) = P(case1) + P(case2) = 1 - r_{s,j^*}^2(t) \quad (14)$$

Similarly, the probability that the offspring $x \in s$ is derived from $p^{j^*}(x,t)$ is,

$$P\left(x \sim p^{j^*}(x,t) | x \in s\right) = r_{s,j^*}^2(t). \quad (15)$$

According to Algorithm 2, the inter-transformation relationship $r_{s,j}(t)$ can be calculated by the degree of overlap between two seed sets,

$$r_{s,j}(t) = \frac{\sum_{m=1}^{N}\sum_{n=1}^{N} I\left(x_s^m(t), x_j^n(t)\right)}{N \times N \times K}, \quad (16)$$

where $x_s^i(t)$ is the $i$-th seed set in the $s$-th sub-population at step $t$. $I(x_s^m(t), x_j^n(t))$ is the number of identical seeds in $x_s^m(t)$ and $x_j^n(t)$. Combining Eq. (14), Eq. (15), and Eq. (16), the *mixed probability distribution* $p^s(x,t)$ can be obtained.

Furthermore, an asymptotic global convergence analysis of MTEFIM is presented.

**Assumption 1.** *It is assumed that: 1) the population size is very large ($N \to \infty$), 2) a parent-centered genetic operator and truncation selection with truncation parameter $\alpha$ are employed, 3) $p^s(x,0)$ is positive and continuous in search space X, s=1, …, S, 4) truncation parameter* $\alpha < 1 - r_{s,j^*}^2(t)$.

In the proof, a more complex truncation selection is considered instead of elitism because evolutionary algorithms with the elitism strategy trivially have asymptotic global convergence properties.

**Theorem 1.** *Assuming that **Assumption 1** is satisfied, then MTEFIM has asymptotic global convergence on each transformation in the multi-transformation environment,*

$$\lim_{t \to \infty} E[\sigma_s(x)] = \int_X \sigma_s(x) \cdot p^s(x,t) \cdot dx = \sigma_s^*, s=1,...,S. \quad (16)$$

TABLE II
COMPARISON OF THE STATISTICAL RESULTS OF THE INFLUENCE SPREAD (AVERAGE ± STANDARD) OBTAINED BY MTEFIM AND MTEFIM-NK OVER 20 INDEPENDENT RUNS.

| Networks | k | MTEFIM | MTEFIM-NK |
|---|---|---|---|
| GN-Network | 3 | **10.396**±0.044 | 9.224±0.206(−) |
| | 12 | **34.209**±0.105 | 32.043±0.117(−) |
| | 21 | **50.995**±0.193 | 48.658±0.286(−) |
| | 30 | **63.596**±0.161 | 61.843±0.572(−) |
| Email URIV | 3 | **46.963**±0.373 | 45.476±0.452(−) |
| | 12 | **94.714**±0.000 | **94.714**±0.683(≈) |
| | 21 | **124.07**±0.810 | 122.769±1.242(−) |
| | 30 | **146.776**±0.574 | 145.341±0.729(−) |
| Hamsterster | 3 | **202.382**±0.021 | 202.336±0.025(≈) |
| | 12 | **231.653**±0.000 | **231.653**±0.000(≈) |
| | 21 | **255.163**±1.043 | 252.619±1.143(−) |
| | 30 | **270.074**±0.962 | 267.961±1.549(−) |
| Ego-facebook | 3 | **868.303**±0.000 | **868.303**±0.000(≈) |
| | 12 | **923.177**±8.409 | 889.976±6.652(−) |
| | 21 | **932.373**±4.190 | 923.004±0.179(−) |
| | 30 | **949.309**±10.298 | 924.221±3.575(−) |
| Fb-pages-public-figure | 3 | **1168.377**±0.906 | 1165.727±1.112(−) |
| | 12 | **1183.986**±3.894 | 1170.890±1.143(−) |
| | 21 | **1203.110**±4.854 | 1178.063±0.863(−) |
| | 30 | **1225.232**±7.250 | 1181.378±4.460(−) |
| NetHEPT | 3 | **45.411**±0.001 | 40.006±1.402(−) |
| | 12 | **116.493**±0.144 | 102.909±0.250(−) |
| | 21 | **158.683**±0.944 | 154.892±1.833(−) |
| | 30 | **194.212**±0.862 | 187.453±2.202(−) |
| −/≈/+ | | − | 0/4/20 |

*Proof.* The proof is in Appendix A, which is similar to a theoretical analysis of most estimations of the distribution algorithms [53] and MFEA [36]. This theorem states that, as the number of iterations increases, the population distribution of each transformation moves towards the position of the global optimum in the search space.

## IV. EXPERIMENTAL STUDY

In this section, an analysis of the knowledge transfer process, inter-transformation relationship estimation, and selection of the output seed set in MTEFIM are first given. Then, the efficacy of the proposed MTEFIM on a range of networks is presented. For the case of synthetic benchmarks and real-world networks, several methods are adopted as baselines for comparison in terms of the influence spread and the running time. The Monte Carlo spread simulation with 10000 runs is employed to evaluate the influence spread of the seed sets obtained by all methods. All the experiments are performed on a PC with a 3.00-GHz Intel Core i7-9700 CPU and 16-GB RAM, and source codes are implemented in Python. The following observations are explained in this section:

1) The knowledge transfer strategy in MTEFIM can improve performance by leveraging common knowledge across transformations (see Section IV.B).

2) For the IM problem, the knowledge transfer mechanism in MTEFIM outperforms those in other EMTO methods (see Section IV.B).

3) Parameter $r$ can explicitly reflect the relationship across the transformations (see Section IV.C).

4) The output seed set selection strategy, SOSS, is effective

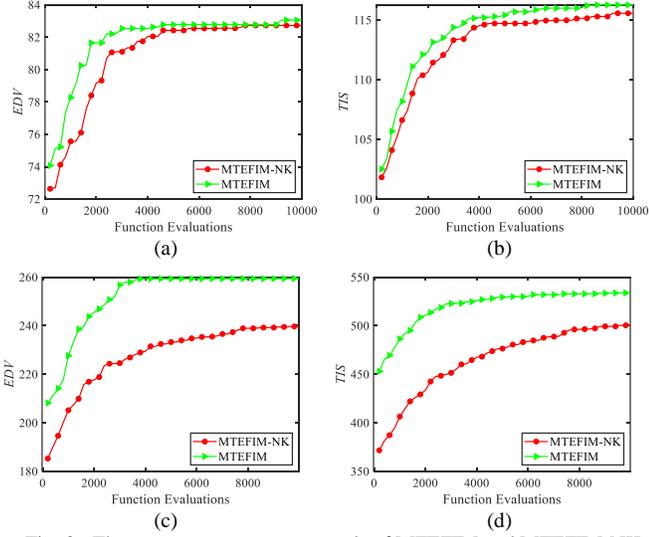

Fig. 8. The average convergence trends of MTEFIM and MTEFIM-NK on two representative networks with $k = 30$ over 20 independent runs, where the $x$-axis represents the value of EDV(TIS), and the $y$-axis represents the function evaluations. (a) Email URIV-EDV, (b) Email URIV-TIS, (c) Fb-pages-public-figure-EDV, and (d) Fb-pages-public-figure-TIS.

(see Section IV.D).

5) The combination of EDV and TIS provides the best performance in a multi-transformation environment, which illustrates the rationale for choosing EDV and TIS (see Section IV.E).

6) Compared with popular IM methods, MTEFIM demonstrates a highly competitive performance by utilizing unique knowledge of different transformations (see Section IV.F).

A. *Experiment Setting*

1) **Datasets**

The performance of MTEFIM is validated on a well-known synthetic network and five real-world networks [54]-[57], including GN-Network, Email URIV, Hamsterster, Ego-facebook, Fb-pages-public-figure, and NetHEPT[2]. The GN networks are employed to represent modular networks with 4 communities and 128 nodes, where each node has the same degree of 16. The parameter $\mu$ is used to control the number of links between communities. All baselines are tested on a GN benchmark network with $\mu=1$ in this article. Email URIV is the category of email networks. The Hamsterster network represents the friendships and family connections between members of the website http://www.hamsterster.com. The Ego-facebook network consists of friends lists from survey participants using Facebook. The large-scale Fb-pages-public-figure network describes Facebook page networks of different categories. The large-scale NetHEPT is a collaborative network of paper co-authors. The basic characteristics of these networks are shown in Table I, where $|V|$ and $|E|$ denote the number of nodes and edges, respectively. *AC* is the average clustering coefficient. The IC model, as a widely popular influence spread model, is employed in this article. According to $|V|$ and *AC*, the configuration of the activation probability $p$ in the IC model is listed in Table I. In addition, EDV and TIS, two popular proxy models, are selected as alternative transformations of IM problems.

[2] https://networkrepository.com

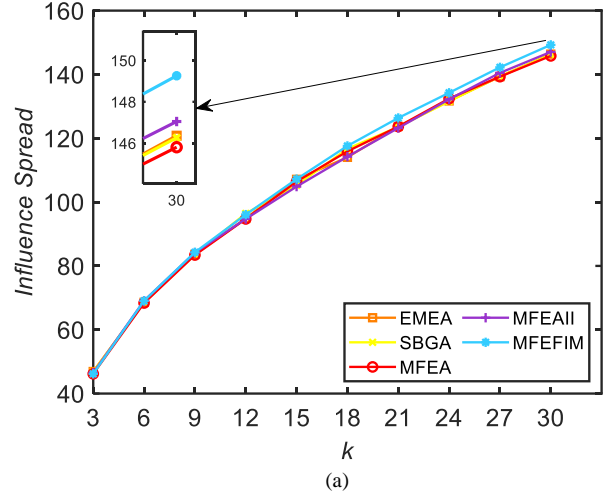

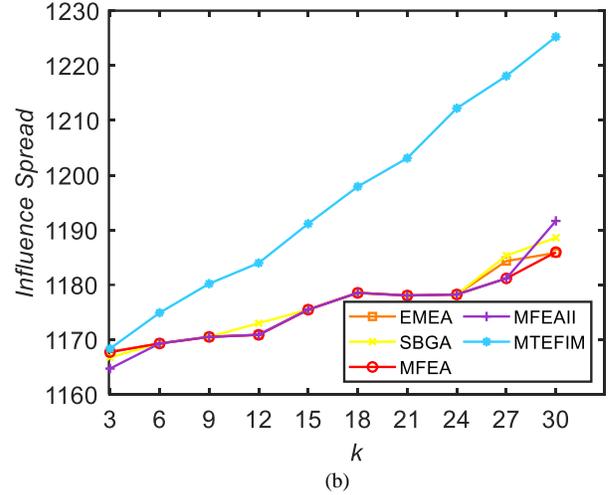

Fig. 9. Influence spread obtained by different EMTO algorithms on the Email URIV and the Fb-pages-public-figure network. (a) Email URIV network, and (b) Fb-pages-public-figure network.

2) **Methods**

To demonstrate the effectiveness of knowledge transfer in MTEFIM, MTEFIM-NK, a variant version of MTEFIM, is constructed by removing the knowledge transfer process (Lines 10~13 in Algorithm 1). In addition, four popular EMTO methods are regarded as comparison algorithms to illustrate the effectiveness of the proposed knowledge transfer strategy, including MFEA [33], MFEAII [36], EMEA [34], and SBEA [35]. MFEA, the first one tailored for the MTO problem, has been applied to many practical problems [58]-[61]. MFEAII tries to minimize the negative transfer tendency across tasks, to improve knowledge utilization in MFEA. EMEA, an explicit EMTO algorithm, performs individual transfers across tasks using a denoising autoencoder. SBEA is a novel many-task optimization algorithm inspired by the symbiosis in biocoenosis. All methods use the same genetic operators and SOSS proposed in this article for a fair comparison. The population size in MFEA and MFEAII is 200, and the population size of each transformation in the other methods is set to 100. The probabilities of crossover and mutation are accordingly set to 1.0 and $1/k$. The maximum number of function evaluations in all methods is set to $5000 \times S$, where $S$ is the number of transformations.

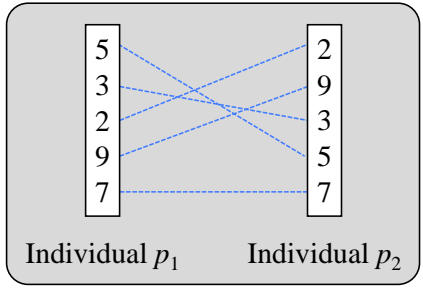

Fig. 10. An illustration of two individuals $p_1$ and $p_2$ represent the same seed set.

Eight popular methods for the IM problem are selected as baselines to verify the performance of MTEFIM, including Degree [62], PageRank [63], SDD [10], CELF++ [14], EDVEA [18], TISEA [19], MA-IM [20], and EDRL-LM [26]. Degree, PageRank, and SDD are three popular heuristic methods. These methods sort nodes according to different indicators and then select the top $k$ nodes as the seed set. CELF++ is an improved greedy algorithm that exploits the sub-modularity of the objective function. The number of Monte-Carlo simulations is set to 10000 in this method. Three popular genetic operator-based EAs are taken as comparison algorithms. The optimization objectives of EDVEA and TISEA are accordingly EDV and TIS. The same genetic operators are employed in these methods for a fair comparison. MA-IM is a popular memetic algorithm for the IM problem that contains two major components, a problem-specific population initialization and a local search based on similarity. EDRL-LM is a state-of-the-art evolutionary deep reinforcement learning framework for the IM to optimize the 2-hop influence spread. The other parameters of all comparison methods are consistent with those of the original papers.

### B. Analysis of Knowledge Transfer Process

Table II lists the influence spread obtained by MTEFIM and MTEFIM-NK over 20 independent runs, where the size of the selected seed set $k$ is set to 3, 12, 21, and 30. The symbols "≈", "+", and "−" indicate that MTEFIM-NK is similar, significantly better, and worse than MTEFIM on the Wilcoxon rank-sum test with a confidence level of 0.95 [64]. The best values are indicated in boldface.

For small-scale social networks such as the GN-Network, Email ERIV, and Hamsterster, both MTEFIM-NK and MTEFIM can find a good seed set. For larger social networks such as Ego-facebook, Fb-pages-public-figure, and NetHEPT, MTEFIM outperforms MTEFIM-NK. In addition, the performance gap across MTEFIM-NK and MTEFIM becomes larger as $k$ or $N$ increases. These phenomena indicate that the effectiveness of knowledge transfer becomes increasingly evident as the complex network optimization difficulty increases, e.g., the number of network nodes and the dimensionality of the search space. In general, compared with MTEFIM-NK, MTEFIM performs better on 12, and ties 4 out of 16 cases in terms of the influence spread, which shows the effectiveness of the knowledge transfer process in MTEFIM. By transferring individuals across the EDV and the TIS, common knowledge is utilized during the evolutionary process.

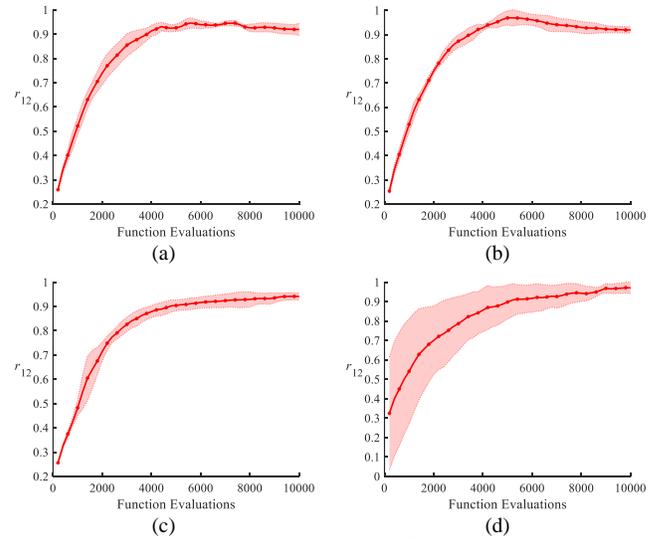

Fig. 11. An illustration of the parameter $r_{12}$ in MTEFIM on four representative real-world social networks, where $k$ is set to 30. (a) Email URIV, (b) Hamsterster, (c) Ego-facebook, and (d) Fb-pages-public-figure.

The average convergence trends of MTEFIM and MTEFIM-NK on two representative networks are shown in Fig. 8 to further illustrate the effectiveness of the knowledge transfer process. As can be seen from Fig. 8, MTEFIM is superior overall to MTEFIM-NK in the average convergence trend of two transformations, EDV and TIS. Compared with MTEFIM-NK, MTEFIM converges faster within 10000 function evaluations. This phenomenon is obvious on a more complex Fb-pages-public-figure network. Based on the above observations, it can be inferred that the knowledge transfer process in MTEFIM can significantly enhance the convergence performance.

In addition, to verify the efficiency of the proposed knowledge transfer process, MTEFIM is compared with MFEA, MFEAII, EMEA, and SBGA on two representative networks, the small-scale Email URIV network and the large-scale Fb-pages-public-figure network, as shown in Fig. 9. In these figures, the x-axis represents the size of seed set $k$, and the y-axis represents the value of the influence spread.

Overall, MTEFIM shows a consistently superior performance. The influence spread obtained by all methods steadily increases as $k$ increases. First, all methods yield similar results in terms of the increasing range in the Email URIV network. This may be because it is not difficult to determine the seed set on a small-scale network. Therefore, the seed set obtained by different methods does not show a clear distinction. Then, in the large-scale Fb-pages-public-figure network, the performance gap of different methods is obvious. The efficiency of the knowledge transfer process in MTEFIM is confirmed since all EMTO methods have the same basic genetic operations and selecting output seed set strategy except for knowledge transfer. Because knowledge transfer in MFEA is purposeless, MFEA cannot adaptively transfer shared knowledge across transformations. In the seed set encoding strategy adopted in this article, multiple individuals may represent the same seed set, as shown in Fig. 10. This phenomenon may lead to the inefficiency of the knowledge transfer process in MFEAII, EMEA, and SBGA. Taking the example of MFEAII, different individuals representing the same seed set may affect the accurate representation of the probability distribution of the current task, resulting in the model inaccuracy. In the knowledge transfer strategy, the degree of overlap between seed sets is estimated to capture the relationship between the transformations, thus avoiding this issue.

TABLE III
THE PROBABILITY THAT THE TWO OUTPUT SEED SETS OBTAINED BY MTEFIM WITH MCSS AND SOSS ON FOUR REPRESENTATIVE NETWORKS ARE THE SAME OVER 20 INDEPENDENT RUNS.

| Networks | k | Probability |
|---|---|---|
| Email URIV | 3 | 100% |
| | 12 | 100% |
| | 21 | 100% |
| | 30 | 100% |
| Hamsterster | 3 | 100% |
| | 12 | 100% |
| | 21 | 100% |
| | 30 | 100% |
| Ego-facebook | 3 | 100% |
| | 12 | 100% |
| | 21 | 100% |
| | 30 | 100% |
| Fb-pages-public-figure | 3 | 90% |
| | 12 | 90% |
| | 21 | 75% |
| | 30 | 75% |

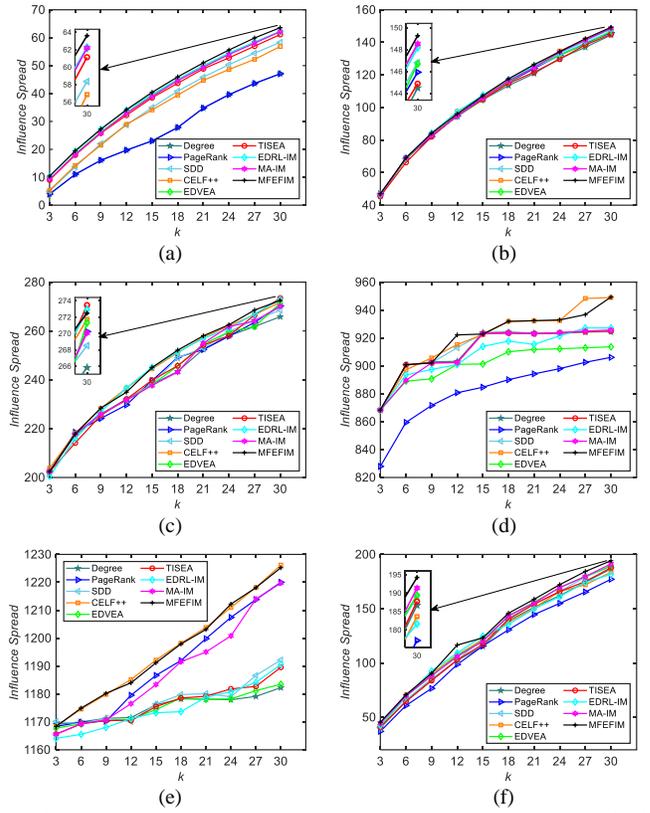

## C. Analysis of Inter-transformation Relationship Estimation

The inter-transformation relationship $R$ is central to controlling the degree of knowledge transfer across transformations in MTEFIM. By calculating the degree of overlap across populations online, the knowledge exchange between transformations is adaptive during the evolutionary process. Fig. 11 shows an illustration of the parameter $r_{12}$ in MTEFIM on four representative social networks, where $k$ is set to 30. In these figures, the x-axis represents function evaluations, and the y-axis refers to the parameter $r_{12}$.

These curves highlight the synergies between transformation pairs of the EDV and the TIS. As shown in Fig. 11, a high value of $r_{12}$ is estimated on all networks, showing high similarities across the EDV and the TIS. This is consistent with the conclusion of the similarity analysis between fitness landscapes (see Fig. 1). Therefore, based on the observations, the inter-transformation relationship estimation can capture potential synergies between transformations. In addition, the number of transferred individuals is determined by $r_{12}$. In different stages of evolution, the estimated $r_{12}$ adaptively guides different degrees of knowledge exchange to accelerate the MTEFIM convergence.

Fig. 12. Influence spread obtained by different methods for IM on both the GN benchmark network and real-world social networks. (a) GN Network, (b) Email URIV, (c) Hamsterster, (d) Ego-facebook, (e) Fb-pages-public-figure, and (f) NetHEPT.

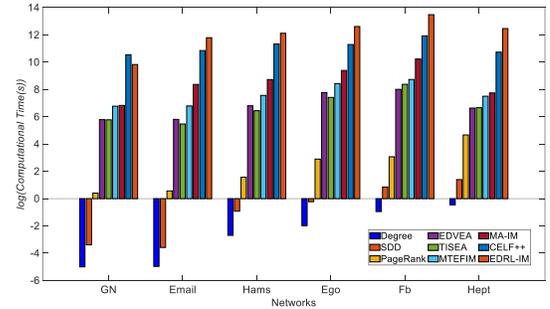

Fig. 13. Runtime analysis of the compared methods on both the GN benchmark network and real-world social networks.

## D. Analysis of Selecting Output Seed Set

After obtaining the optimal individual of each transformation, the comprehensive performance of the individuals in all transformations is considered to select the final output seed set. Table III lists the probability that the two output seed sets obtained by MTEFIM with MCSS and SOSS on four representative networks are the same over 20 independent runs, where the user's prior preference $C$ is set to $1/S$ to treat each transformation equally.

It can be seen from Table III that MCSS and SOSS select the same output seed set in most cases, which indicates the effectiveness of the proposed output seed set selection strategy. It is worth noting that in the large-scale Fb-pages-public-figure network, the individual with the best overall performance among all transformations is not necessarily the best in the Monte Carlo simulation. This may be because there is some preference knowledge in the search process of different transformations, resulting in optimal individual deviations. However, the proposed method provides an efficient output seed set selection strategy that avoids expensive Monte Carlo simulations.

TABLE IV
COMPARISON OF THE STATISTICAL RESULTS OF THE INFLUENCE SPREAD (AVERAGE ± STANDARD) OBTAINED BY DIFFERENT TRANSFORMATIONS OVER 20 INDEPENDENT RUNS.

| Networks | $k$ | EDV | TIS | PS | EDV&TIS | EDV&PS | TIS&PS | EDV&TIS&PS |
|---|---|---|---|---|---|---|---|---|
| GN-Network | 3 | 9.162±0.074 | 9.004±0.212 | 9.742±0.263 | 10.396±0.044 | 10.435±0.058 | 10.396±0.166 | 10.297±0.141 |
|  | 12 | 33.041±0.065 | 32.220±0.137 | 32.056±0.663 | 34.209±0.105 | 34.266±0.889 | 34.313±0.377 | 34.116±0.139 |
|  | 21 | 49.610±0.125 | 48.845±0.256 | 47.266±1.055 | 50.995±0.193 | 50.991±0.120 | 50.440±0.212 | 50.996±0.399 |
|  | 30 | 62.162±0.135 | 61.129±0.254 | 60.102±0.841 | 63.596±0.161 | 63.679±0.086 | 63.140±0.334 | 63.499±0.151 |
| NetHEPT | 3 | 41.497±1.510 | 40.872±1.402 | 37.534±4.511 | 45.411±0.001 | 45.498±0.001 | 47.327±1.214 | 45.498±0.924 |
|  | 12 | 107.021±0.000 | 102.878±5.656 | 107.342±3.042 | 116.493±0.144 | 116.468±4.318 | 115.801±3.932 | 116.558±0.137 |
|  | 21 | 156.038±5.480 | 154.225±1.267 | 139.791±4.677 | 158.683±0.944 | 159.404±0.353 | 159.933±2.316 | 159.668±1.043 |
|  | 30 | 189.374±0.609 | 187.714±1.553 | 165.273±4.759 | 194.212±0.862 | 194.043±0.153 | 191.938±0.923 | 193.819±1.168 |

*E. Analysis of Selecting Transformations*

In this section, the performance of MTEFIM with different transformations is further demonstrated. In addition to EDV and TIS, another commonly used transformation, probability sorting (PS), is considered, which is a typical diffusion model reduction proxy model [29]. PS computes a simplified representative instance of the studied social network. A single diffusion simulation is performed on PS to approximate the diffusion process on the original social network. Table IV lists the influence spread obtained by different combinations of transformations on the small-scale GN network and the large-scale NetHEPT network over 20 independent runs, where the size of the selected seed set $k$ is set to 3, 12, 21, and 30. Compared to optimizing each transformation independently, optimizing any combination of transformations has better performance in terms of the influence spread, in which useful knowledge carried by transferred individuals can be efficiently shared. Furthermore, different combinations of transformations have a similar performance, which also shows that the proposed method can cope with scenarios with different proxy models rather than being tailored for a certain one.

*F. Performance Comparison*

To further verify MTEFIM's effectiveness, several comparative experiments are conducted in terms of the influence spread and computational time. Fig. 12 reports the influence spread of various methods for IM on both the GN benchmark network and real-world social networks.

The experimental results in Fig. 12 (a) show that MTEFIM has a better performance than the heuristic methods (Degree, PageRank, and SDD) in the GN networks, an observation that is also prevalent in the real-world networks (see Fig. 12 (b)~(f)). As the size of seed set $k$ (search space) increases, the heuristic method is more likely to be trapped in local optima. This is because they focus on improving the ability to exploit the landscape, while ignoring the ability to explore the landscape. As shown in Fig. 12 (b)~(f), different methods for IM have different performances on real-world networks. In Fig. 12 (b), the differences in the influence spread of nine algorithms are not obvious for the email network. With the increase of $k$, the influence obtained by all methods has steadily extended. Compared with the single-transformation optimization method, EDVEA, and TISEA, MTEFIM produced better results. This implies that MTEFIM can simultaneously utilize all transformations of unique knowledge in the search process to significantly improve performance. This observation is even more evident in large-scale social networks (see Fig. 12 (e)). In Fig. 12 (b) and (c), for the Hamsterster and the Ego-facebook, CELF++, EDRL-LM, and MTEFIM achieve similar results in terms of the influence spread. When $k > 15$, the performances of MTEFIM, EDRL-LM, and CELF++ are better than those of all other methods. In Fig. 13 (e), for the large-scale Fb-pages-public-figure network, MTEFIM is similar to CELF++. The influence spreads of Degree and EDVEA increase slowly with $k$. This might be because these methods ignore overlapping second-degree neighborhoods. In Fig. 12 (f), for the NetHEPT network, MTEFIM outperforms all the other baseline methods, which further illustrates the effectiveness of MTEFIM on large-scale networks. Overall, the proposed MTEFIM has obtained highly competitive results on social networks of different sizes.

GN networks have a clear community structure without obvious heterogeneous structural information. Compared with GN networks, real networks present more heterogeneous structural information. Fig. 12 shows that MTEFIM outperforms greedy methods (CELF++) across the board in synthetic GN networks. In most real-world networks, MTEFIM has a highly competitive performance compared to CELF++. This is because the main greedy principle of CELF++ is tailored based on the heterogeneous information of the network, while MTEFIM applied in this article does not depend on specific network structural properties. In addition, real-world networks show the scale-free property of nodes, which makes it possible to obtain a good influence spread through a small number of nodes with high node degrees.

Fig. 13 shows a runtime analysis of the comparative methods on both the GN benchmark network and the real-world social networks, where computational time refers to the sum of the runtime of a method in ten configurations with $k$ =3, 6, 9, …, 30. First, since three heuristic methods, Degree, SDD, and PageRank, only need to sort nodes according to the indicator and then select the top $k$ nodes as the seed set, their computational cost is extremely meager. In addition, the running time of PageRank increases rapidly with the scale of the network. Second, the single-transformation optimization methods, EDVEA and TISEA, perform better than MTEFIM in terms of computational cost. However, MTEFIM can find a higher quality seed set by optimizing EDV and TIS at the same time. Third, the computational cost of MA-IM and EDRL-LM is higher than that of the other evolution-based methods because the local search in MA-IM and the model learning in EDRL-LM are computationally expensive. Finally, the proposed MTEFIM achieves a highly competitive performance with CELF++ at a far lower computational cost.

## V. CONCLUSION

Inspired by the similarities and unique advantages of different transformations for influence maximization in social networks, a novel multi-transformation evolutionary framework is introduced in this article. By considering the degree of overlap across seed sets of different transformations, the proposed method can capture the inter-transformation potential relationship, which controls the frequency and degree of the knowledge transfer process. In addition, an output seed set selection strategy is provided to avoid manual selection by users. Several experiments on benchmarks and real-world social networks are employed to illustrate the reasonability of the above claims. Compared with EMTO and the IM-specific methods, MTEFIM can obtain a high-quality seed set with a lower computational cost.

Because of the broad application potential of IM, the following topics deserve further research. First, it is still worthwhile to theoretically explore the relationships across IM transformations. Second, a series of swarm intelligence-based methods for IM have shown extraordinary performances [21]-[22], [24]-[25], [46]-[48]. The knowledge transfer process design across multiple transformations in swarm intelligence optimization methods to enhance performance for IM can be taken into consideration. Finally, applying MTEFIM to solve more complex IM problems such as the competitive IM [28] and the multi-round IM [65]-[66] also serves as a promising research topic.

## APPENDIX

### A. Proof of Theorem 1.

*Proof.* The basic idea of truncation selection is that the top $\alpha(t)$ individuals in the fitness ranking of the current population are selected as the next generation population, which can be modeled as,

$$p^s(x,t+1) = \begin{cases} \dfrac{q^s(x,t)}{\alpha(t+1)}, & \text{if } \sigma_s(x) \geq \beta_s(t+1) \\ 0, & \text{if } \sigma_s(x) < \beta_s(t+1) \end{cases}, \quad (1)$$

where $\alpha(t+1) = \int_{\sigma_s(x) \geq \beta_s(t+1)} q^s(x,t) \cdot dx$ is a real number. Since the following formula holds,

$$q^s(x,t) = \left[1 - r_{s,j^*}^2(t)\right] \cdot p^s(x,t) + r_{s,j^*}^2(t) \cdot p^{j^*}(x,t)$$

$$j^* = \arg\max_{j \in \{1,2,\ldots,s\}, j \neq s} r_{s,j}(t) = \dfrac{\sum_{m=1}^{N}\sum_{n=1}^{N} I\left(x_s^m(t), x_j^n(t)\right)}{N \times N \times K} \quad (2)$$

we have,

$$\int_{\sigma_s(x) \geq \beta_s(t)} q^s(x,t) \cdot dx \geq \int_{\sigma_s(x) \geq \beta_s(t)} \left[1 - r_{s,j^*}^2(t)\right] \cdot p^s(x,t) \cdot dx. \quad (3)$$

Because $\int_{\sigma_s(x) \geq \beta_s(t)} p^s(x,t) \cdot dx = 1$, we have,

$$1 - r_{s,j^*}^2(t) \leq \int_{\sigma_s(x) \geq \beta_s(t)} q^s(x,t) \cdot dx. \quad (4)$$

According to **Assumptions 1** *4)*, the following inequality holds,

$$\alpha(t+1) < 1 - r_{s,j^*}^2(t). \quad (5)$$

Then,

$$\int_{\sigma_s(x) \geq \beta_s(t+1)} q^s(x,t) \cdot dx < \int_{\sigma_s(x) \geq \beta_s(t)} q^s(x,t) \cdot dx, \quad (6)$$

i.e. $\beta_s(t) < \beta_s(t+1)$, which implies that there exists a limit,

$$\lim_{t \to \infty} \beta_s(t) = \sigma_s'. \quad (7)$$

Next, the contradiction is employed to prove that $\sigma_s' = \sigma_s^*$. Assume that $\sigma_s' < \sigma_s^*$. According to Eq. (1) and Eq. (2), we have,

$$p^s(x,t) \geq p^s(x,0) \left(\dfrac{1 - r_{s,j^*}^2(t)}{\alpha(t)}\right)^t, \forall x : \sigma_s(x) > \sigma_s'. \quad (8)$$

Because $p^s(x,0)$ is positive and continuous in search space $X$ (**Assumptions 1** *3)*), we obtain,

$$\lim_{t \to \infty} p^s(x,t) = +\infty, \forall x : \sigma_s(x) > \sigma_s'. \quad (9)$$

Let $X' = \{x | x \in X, \sigma_s(x) > \sigma_s'\}$. According to Fatou's lemma, we obtain,

$$\lim_{t \to \infty} \int_{X'} p^s(x,t) \cdot dx = +\infty. \quad (10)$$

Because Eq. (10) contradicts the fact that $p^s(x,t)$ is a probability density function, we have,

$$\lim_{t \to \infty} \beta_s(t) = \sigma_s^*. \quad (11)$$

Due to $E[\sigma_s(x)] = \int_X \sigma_s(x) \cdot p^s(x,t) \cdot dx \geq \beta_s(t)$, we have,

$$\lim_{t \to \infty} E[\sigma_s(x)] = \int_X \sigma_s(x) \cdot p^s(x,t) \cdot dx = \sigma_s^*. \quad (12)$$